\documentclass{article}

\usepackage{amsmath}
\usepackage{nicefrac}
\usepackage{graphicx}
\usepackage{caption}
\usepackage{wrapfig}
\usepackage{multirow}
\usepackage{colortbl}
\usepackage[most]{tcolorbox}
\usepackage{enumitem}

\DeclareMathOperator*{\expec}{\mathbb{E}}

\definecolor{lightblue}{rgb}{0.85,0.92,1}

   \usepackage[nonatbib,final]{neurips_2025}

\usepackage[utf8]{inputenc} % allow utf-8 input
\usepackage[T1]{fontenc}    % use 8-bit T1 fonts
\usepackage{hyperref}       % hyperlinks
\usepackage{url}            % simple URL typesetting
\usepackage{booktabs}       % professional-quality tables
\usepackage{amsfonts}       % blackboard math symbols
\usepackage{nicefrac}       % compact symbols for 1/2, etc.
\usepackage{microtype}      % microtypography
\usepackage{xcolor}         % colors
\usepackage{natbib}
\usepackage{amssymb}
\usepackage{amsthm}

\newtheorem{theorem}{Theorem}

\title{Stop Summation: Min-Form Credit Assignment Is All Process Reward Model Needs for Reasoning}

\author{Jie Cheng$^{1,2}$,
  Gang Xiong$^{1,2}$,
  Ruixi Qiao$^{1,2}$,
  Lijun Li$^{3}$,
  Chao Guo$^{1}$,\\
  \textbf{Junle Wang}$^{4}$, 
  \textbf{Yisheng Lv}$^{1,2,5}$\thanks{Corresponding author}~,
  \textbf{Fei-Yue Wang}$^{5,6,7,8}$\\
  $^{1}$State Key Laboratory of Multimodal Artificial Intelligence Systems, \\Institute of Automation, Chinese Academy of Sciences\\ %\\  ~~Institute of Automation, Chinese Academy of Sciences\\
  $^{2}$School of Artificial Intelligence, University of Chinese Academy of Sciences\\
  $^{3}$Shanghai Artificial Intelligence Laboratory~~$^{4}$Tencent\\
  $^{5}$Faculty of Innovation Engineering, Macau University of Science and Technology\\
  $^{6}$DeSci Center of Parallel Intelligence, Obuda University\\
  $^{7}$Research Center for Chinese Economics and Social Security, \\University of Chinese Academy of Sciences\\
  $^{8}$Institute of Automation, Chinese Academy of Sciences
}

\begin{document}

\maketitle

\begin{abstract}
Process reward models (PRMs) have proven effective for test-time scaling of Large Language Models (LLMs) on challenging reasoning tasks. However, reward hacking issues with PRMs limit their successful application in reinforcement fine-tuning. In this paper, we identify the main cause of PRM-induced reward hacking: the canonical summation-form credit assignment in reinforcement learning (RL), which defines the value as cumulative gamma-decayed future rewards, easily induces LLMs to hack steps with high rewards. To address this, we propose PURE: \underline{\textbf{P}}rocess s\underline{\textbf{U}}pervised \underline{\textbf{R}}einforcement l\underline{\textbf{E}}arning. The key innovation of PURE is a min-form credit assignment that formulates the value function as the minimum of future rewards. This method significantly alleviates reward hacking by limiting the value function range and distributing advantages more reasonably. Through extensive experiments on 3 base models, we show that PRM-based approaches enabling min-form credit assignment achieve comparable reasoning performance to verifiable reward-based methods within only 30\% steps. In contrast, the canonical sum-form credit assignment collapses training even at the beginning! Additionally, when we supplement PRM-based fine-tuning with just 10\% verifiable rewards, we further alleviate reward hacking and produce the best fine-tuned model based on Qwen2.5-Math-7B in our experiments, achieving 82.5\% accuracy on AMC23 and 53.3\% average accuracy across 5 benchmarks. Moreover, we summarize the observed reward hacking cases and analyze the causes of training collapse. We release our code and model weights at \url{https://github.com/CJReinforce/PURE}.
\end{abstract}
\section{Introduction} \label{sec:intro}
Reinforcement fine-tuning (RFT) of large language models (LLMs) for reasoning tasks has shown promise in developing advanced problem-solving abilities. Recent advancements~\citep{wang2025step,zeng2025glm}, such as DeepSeek R1-Zero~\citep{guo2025deepseek} and Kimi K1.5~\citep{team2025kimi}, have demonstrated strong reasoning skills through RFT with verifiable rewards, which provide sparse feedback for the entire response. However, as response length increases, sparse rewards potentially lead to inefficient learning~\citep{sutton2018reinforcement,andrychowicz2017hindsight}. 

In contrast, process reward models (PRMs) offer dense feedback at each step of a response. PRMs have proven effective in improving LLMs’ performance on challenging reasoning tasks through test-time scaling~\citep{lightman2023let,wang2023math,guan2025rstar}. However, successful applications of PRM in RFT for LLMs remains limited~\citep{setlur2024rewarding}. A key challenge of PRMs is that the neutral network-generated rewards can lead to reward hacking during training~\citep{weng2024rewardhack,guo2025deepseek}, causing unintended optimization toward higher rewards. Recent studies~\citep{yuan2024free,cui2025process} have explored implicit PRMs for fine-tuning LLMs, which resembles DPO-style reward formulation~\citep{rafailov2023direct}, but these still heavily relies on ground-truth verifiable rewards to provide training signals for the system. The question of \textit{what causes reward hacking in PRM-based RFT and how to address it effectively} has not been explored widely.

To answer this question, we examine the usage of PRM in test-time scaling and identify a mismatch between test-time and training-time objectives. Thus in this paper, we propose PURE: \underline{\textbf{P}}rocess s\underline{\textbf{U}}pervised \underline{\textbf{R}}einforcement l\underline{\textbf{E}}arning, which introduces a min-form credit assignment method to align these objectives. Through analysis and experiments, we find that the canonical formulation of credit assignment in RL, named summation-form credit assignment that defines value as cumulative gamma-decayed future rewards, easily induces LLMs to hack high-reward steps and collapse training. Instead, we use the minimum future reward to quantify credit assignment, which constrains the value function’s range and assigns advantages more reasonably to stabilize RL training (see the analysis in \S~\ref{sec:min-form credit assignment}). PURE offers several benefits: (1) it is simple to implement, requiring only a transformation of process rewards without additional code changes; (2) it achieves comparable or better performance in PRM-based RFT compared to recent R1-replication studies with around 3$\times$ efficiency gains; and (3) it supports the integration of both dense process rewards and sparse verifiable rewards. When combining both reward types in the PURE framework, we find that the auxiliary of few ground-truth signals further mitigates reward hacking caused by PRM.

In experiments, we first train a PRM using the PRM800K dataset~\citep{lightman2023let}, a human-annotated dataset that evaluates the correctness of each step. Then we apply PURE framework to 3 models: Qwen2.5-7B, Qwen2.5-Math-7B, and Qwen2.5-Math-1.5B, using 3 reward configurations: only verifiable rewards, only process rewards, and a combination of both. This comprehensive setup allows us to compare PRM-based and verifiable reward-based RFT approaches. We also compare sum-form and min-form credit assignment methods when enabling PRMs and find that the former even collapses training at the beginning. In the discussion, we summarize observed reward hacking cases and analyze the causes of training collapse. Our key findings are outlined below.

\vspace{-2pt}

\begin{tcolorbox}[colback=lightblue!80,breakable]
\begin{enumerate}[leftmargin=1em]
    \item Summation-form credit assignment easily induces LLMs to hack high-reward steps, leading to LLMs that prioritize thinking over problem-solving. Our min-form credit assignment avoids such reward hacking. (\S~\ref{sec:min-form credit assignment})
    \item PRM-based RFT achieves performance comparable to the verifiable reward-based approach with around 3$\times$ efficiency gains if using our min-form credit assignment; otherwise, the training collapses even at the beginning! (\S~\ref{sec:main results})
    \item The auxiliary of few ground-truth signals further mitigates PRM-induced reward hacking, achieving 82.5\% accuracy on AMC23 and 53.3\% average accuracy across MATH-500, Minerva Math, Olympiad Bench, AIME24, and AMC23 when using Qwen2.5-Math-7B as the base model. (\S~\ref{sec:main results})
    \item We identify 3 types of PRM-induced reward hacking: (1) only thinking, not solving, (2) extremely few steps (1 step), (3) extremely few steps (0 step). We analyze causes, show examples, and provide solutions for each. (\S~\ref{sec:reward hacking})
    \item Long, highly repetitive samples that the verifier ruled correct cause training to collapse. These pseudo-positive samples, undetected by verifiers, provide numerous incorrect signals and collapse training all of a sudden (within 5 gradient steps). (\S~\ref{sec:cause of training collapse})
\end{enumerate}
\end{tcolorbox}

% Our primary contributions are as follows. First, we find the cause of reward hacking during PRM-based RFT through semi-quantitative analysis and experiments, and propose min-form credit assignment method with the minimal code changes to alleviate the issue. Second, through extensive experiments, we find PRM-based RFT can achieve similar performance to the VR-based approach if using our proposed min-form credit assignment method, otherwise causes training collapse even at the begining!
\section{Preliminaries}
\vspace{-5pt}
\subsection{Credit Assignment Problem in Reinforcement Learning}
\citet{minsky2007steps} described the credit assignment problem as "how to distribute the credit of success among the multitude of decisions involved". In other words, credit assignment is the problem of \textit{estimating the influence of an action} over an outcome from experience. Various assignment functions have been proposed to quantify the influence of actions~\citep{pignatelli2023survey}. Here, we adopt the state-action value as the assignment function to detail the problem~\citep{pmlr-v235-cheng24k,ICLR2025_689cffc9}.

We model LLM reasoning as a step-level Markov Decision Process. Given a prompt, LLM generates steps sequentially. At step $t$, LLM $\pi$ takes the prompt $p$ and previous steps $\{a_1,\cdots,a_{t-1}\}$ as the input state $s_t$, where $s_t=(p,a_1,\cdots,a_{t-1})$. Then LLM generates a step $a_t$, sampled as $a_t\sim \pi(\cdot|s_t)$. PRM $R^p$ emits a process reward $r^p_t=R^p(s_t,a_t)$. The canonical formulation of state-action value is:
\begin{equation}
    Q^\pi(s_t, a_t) = \mathbb{E}\left[\sum_{i\geq t} \gamma^{i-t} r^p_i\right] \label{eq:definition of classical Q}
\end{equation}
where $\gamma$ is the discount factor. Eq. (\ref{eq:definition of classical Q}) shows that the influence of an action decreases over time, weighted by a discount factor $\gamma$. Actions closer to future outcomes have greater influence. Thus, the state-action value quantifies credit assignment. With Eq. (\ref{eq:definition of classical Q}), we can derive the advantage function~\citep{schulman2017proximal,shao2024deepseekmath,ahmadian2024back,hu2025reinforce++} and update LLM using the policy gradient loss.

% \vspace{-5pt}
\subsection{Process Reward vs. Verifiable Reward} \label{sec:pr vs. vr}
Verifiable rewards are sparse, rule-based rewards assigned to an entire response. For a prompt, LLM generates a sequence of steps $\{a_1, \dots, a_n\}$. After the final step $a_n$, a verifiable reward $r^v$ is assigned based on whether the response matches the ground-truth answer. Verifiable rewards provide a straightforward, ground-truth signal for RL training, which has been commonly used in the training pipeline for advanced models~\citep{yue2024sc,guo2025deepseek,team2025kimi}. In contrast, process rewards $\{r^p_1,\cdots,r^p_n\}$ are dense rewards to evaluate the quality of each step. Process rewards offer unique benefits in test-time scaling. During inference, LLM generates multiple responses for a single prompt, and PRM scores each step for each response. Prior work~\citep{lightman2023let,wang2023math,zhang2025lessons} typically aggregates process rewards into a outcome-level score using the minimum value, $\min(r^p_1,\cdots,r^p_n)$, and selects the best response based on this outcome-level score. This PRM-based approach outperforms majority voting and other reward model-based approaches.

Although dense rewards support effective training in traditional RL applications, they struggle to effectively fine-tune LLM for advanced reasoning~\citet{guo2025deepseek}. Therefore, how to make PRMs as effective in training as they are in test-time scaling is an important and not widely studied topic.
% \vspace{-5pt}
\section{PURE: Process Supervised Reinforcement Learning}
To effectively fine-tune LLMs using PRMs, we propose PURE: \underline{\textbf{P}}rocess s\underline{\textbf{U}}pervised \underline{\textbf{R}}einforcement l\underline{\textbf{E}}arning. PURE leverages a novel min-form formulation to quantify the credit assignment, inspired by the test-time application of PRM. In this section, we first detail the min-form credit assignment and analyze its effectiveness in \S~\ref{sec:min-form credit assignment}. Subsequently, we introduce an advantage estimator tailored for process rewards in \S~\ref{sec:advantage estimation}. 

\begin{figure}[th]
    \centering
    \includegraphics[width=0.8\textwidth]{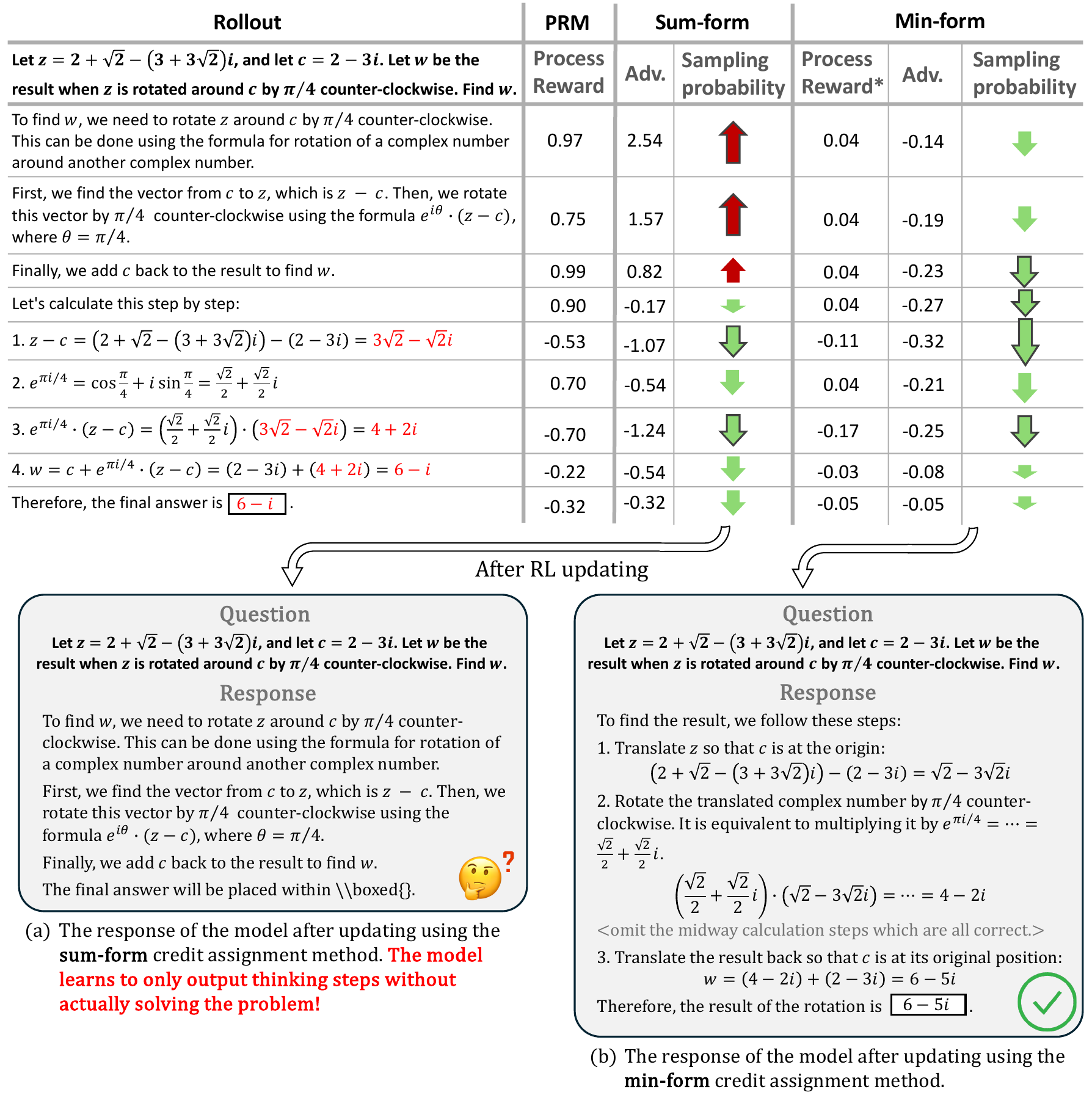}
    \caption{Comparison of summation-form and min-form credit assignment. Adv. and Process reward* in the table means advantage and transformed process reward, respectively. The \textcolor{red}{incorrect steps} in the rollout are highlighted in \textcolor{red}{red}, and our PRM reasonably assigns negative scores to these steps. For simplicity, advantage baseline and KL penalty terms are omitted in advantage calculation here, and discount factor $\gamma$ and transform temperature $T$ are set to 1. Arrows indicate changes in sampling probability, with larger changes marked by contoured arrows.}
    \vspace{-10pt}
    \label{fig:difference of credit assignment}
\end{figure}
\vspace{-5pt}
\subsection{Min-form Credit Assignment} \label{sec:min-form credit assignment}
Consider a prompt $p$ and a LLM $\pi$ parameterized by $\theta$, generating a $n$-step response denoted as $(a_1,\cdots,a_n)$. At step $t$, LLM $\pi_\theta$ takes the prompt $p$ and previous steps $\{a_1,\cdots,a_{t-1}\}$ as the input state $s_t$, where $s_t=(p,a_1,\cdots,a_{t-1})$. Then LLM generates step $a_t$, sampled as $a_t\sim \pi_\theta(\cdot|s_t)$. A PRM $R^p$ emits a process reward $r^p_t=R^p(s_t,a_t)$. Unlike traditional RL whose objective is the summation of discounted rewards
% (like Eq.~\eqref{eq:definition of classical Q})
, the usage of PRM in test-time scaling suggests that RL for reasoning tasks should optimize the following objective:
\begin{equation}
    \arg\max_\theta~~\mathbb{E}\left[\min (r^p_1,\cdots,r^p_n)\right]
\end{equation}
where the expectation is over prompts and step-level actions. This objective implies: \textbf{(i)} Only the ``worst’’ step that gets the minimum process reward determines the value of the entire response. \textbf{(ii)} For steps before the worst step, their existence as the input state induces LLM to generate the worst step. \textbf{(iii)} Steps after the worst step actually do not contribute to the objective. 

Therefore, we define the min-form credit assignment function as follows: For a $n$-step response, let step $w$ be the worst step, \textit{i.e.}, $w=\arg\min(r^p_1,\cdots,r^p_n)$. The return $G$ and state-action value $Q^\pi$ functions are defined as follows:
\begin{align}
    G(s_t,a_t | \tau) &= \begin{cases}
        \min (r^p_t, \cdots, r^p_n), & \text{if } t \leq w \\
        0, & \text{if } t > w
        \end{cases} \label{eq:min-form return} \\ 
    Q^\pi(s_t,a_t) &= \expec_{\tau} \left[ G(s_t,a_t | \tau) \right] \label{eq:min-form Q}
\end{align}
where $\tau=(s_1,a_1,r^p_1,\cdots,s_n,a_n,r^p_n)$ denotes the trajectory. Eq. (\ref{eq:min-form Q}) is a quantitative expression of the above three-fold analysis of the objective. To implement this in the simplest way possible, we transform the process rewards in trajectory $\tau$ using:
\begin{equation}
    r_i^{p*}=\frac{\exp(\nicefrac{-r^p_i}T)}{\sum_{j=1}^n \exp(\nicefrac{-r^p_j}T)}\cdot r^p_i \label{eq:transform reward}
\end{equation}
where $T$ is the transform temperature, and $r_i^{p*}$ is the transformed process reward. The transform function assigns higher weights to lower rewards. After transformation, the rest of code remains unchanged. We explain this as follows. As the transform temperature $T \rightarrow 0^+$, Eq. (\ref{eq:transform reward}) yields $r_w^{p*}=r_w^p$ for the worst step $w$ and $r_i^{p*}=0$ for $i\neq w$. Setting $\gamma=1$, the return becomes: 
\textbf{(i)} For step $t \leq w$,
\begin{equation}
    G_t=\sum_{i=t}^n\gamma^{i-t} \cdot r^{p*}_i= r^p_w = \min (r^p_t, \cdots, r^p_n)
\end{equation}
\resizebox{\textwidth}{!}{
\textbf{(ii)} For $t>w$, $G_t=0$. The return exactly matches Eq.~\eqref{eq:min-form return} without altering the return computation logic.}

\textbf{Simple implementation.} To align with widely adopted token-level PPO loss, we convert step-level rewards to token-level rewards. We exclusively assign the transformed process reward to the final token of each step, with all other tokens receiving a reward of 0. To preserve algorithmic flexibility, we also support verifiable rewards assigned to the last token of the complete response, which can be used alongside process rewards.

% We can easily turn the step-level reward and return into token-level to adapt the commonly used token-level PPO loss. Just assign the transformed process reward on the last token of the corresponding step, and assign the rest of tokens 0 reward, which we named token-level transformed process reward. Then no code else has to be changed. Moreover, to ensure algorithmic flexibility, we allow for the verifiable reward in addition to process rewards, which is assigned at the last token of the entire response.

\subsection{Quantitative Analysis}
Here we analyze that the min-form has a smaller value estimation error than the summation-form credit assignment method. We start by first establishing 2 reasonable assumptions:

\textbf{Assumption 1 (Bounded Process Reward Error):} The Process Reward Model, $R^P$, produces a reward estimate $r_t^p = R^P(\cdot)$ for each reasoning step. It is assumed that there exists a "true" or optimal reward $r_t^*$ for that step, and the estimation error is uniformly bounded by a constant $\epsilon \ge 0$. For any step $t$, it holds that $|r_t^p - r_t^*| \le \epsilon$.

\textbf{Assumption 2 (Bounded Rewards):} The true rewards, and consequently the estimated rewards, are bounded. There exists a maximum possible reward $R_{\max}$ such that $|r_t^*| \le R_{\max}$ and $|r_t^p| \le R_{\max} + \epsilon$ for all steps $t$.

\begin{theorem}[\textbf{Q-Value Estimation Error Bound Comparison}]
Under Assumptions 1 and 2, for any state-action pair $(s_t, a_t)$ and a trajectory $\tau$ with $n$ reasoning steps:
\begin{enumerate}[leftmargin=*]
    \item \textbf{Sum-form Error Bound:} The estimation error for the summation-form Q-value, is bounded by:
    $$
    |Q_{\pi}^{sum}(s_t, a_t) - Q_{\pi}^{sum,*}(s_t, a_t)| \le \sum_{i=t}^{t+n-1} \gamma^{i-t} \epsilon
    $$
    For an infinite horizon ($n \to \infty$), this bound becomes:
    $$
    |Q_{\pi}^{sum}(s_t, a_t) - Q_{\pi}^{sum,*}(s_t, a_t)| \le \frac{\epsilon}{1 - \gamma}
    $$

    \item \textbf{Min-form Error Bound:} The estimation error for the min-form Q-value, derived from Eq. (\ref{eq:min-form Q}), is bounded by:
    $$
    |Q_{\pi}^{min}(s_t, a_t) - Q_{\pi}^{min,*}(s_t, a_t)| \le \epsilon
    $$
\end{enumerate}
\end{theorem}
The proof is provided in Appendix \ref{app:proof}. This theorem shows that the error bound of sum-form method is amplified by a factor dependent on the horizon length and the discount factor $\gamma$, accumulating the error over steps. However, the error bound of min-form method is strictly limited by the single-step reward estimation error $\epsilon$ and does not accumulate with the number of steps.

\subsection{Qualitative Analysis}
% \textbf{Analysis.} 
Min-form credit assignment restricts the range of value function to be the same as that of the reward function, which contributes to stabilizing RL training. In contrast, for the summation-form credit assignment, the range of value is determined by the range of reward and the number of steps, causing excessive value as the number of steps increases and unintended reward hacking.

We now visualize the differences between summation-form and our min-form credit assignment using a real training example. As shown in Figure \ref{fig:difference of credit assignment}, incorrect steps are highlighted in red, with arrows showing the magnitude and direction of changes in sampling probability. Steps with relatively large changes in sampling probability are marked by contoured arrows. Note that the first 3 steps are the thinking, and the subsequent 6 steps are the solution. 

Figure \ref{fig:difference of credit assignment} shows that \textbf{summation-form credit assignment significantly alters sampling probabilities}, increasing the sampling probabilities for thinking steps and decreasing that for incorrect solution steps. However, this results in a shotcut towards reward hacking: the model learns to only output thinking steps without actually solving the problem! In other words, \textbf{the model hacks the implicit pattern inside high-reward steps}, \textit{i.e.}, thinking. In contrast, our min-form credit assignment reduces sampling probabilities across all steps, with the largest reduction at the first incorrect step. This aligns with the behavior of verifiable rewards (an incorrect final answer leads to a reduction in sampling probability of all steps) and assigns advantages more rationally based on step correctness.

\vspace{-5pt}
\subsection{Advantage Estimation} \label{sec:advantage estimation}
Following the token-level rewards in \S~\ref{sec:min-form credit assignment}, we are ready to compute advantages. We compare several advantage estimators, including GAE~\citep{schulman2017proximal}, RLOO~\citep{ahmadian2024back}, GRPO~\citep{shao2024deepseekmath}, and REINFORCE++~\citep{hu2025reinforce++}, as discussed in \S~\ref{sec:other rl algo}. RLOO are strong enough to produce stable and effective results. 

For a single prompt, LLM generates $K$ responses. The maximum generation length is limited to $N$. Let $r^v_i$ and $r^{p*}_{i,j}$ $(j=1,\cdots,N)$ denote the verifiable reward and token-level transformed process rewards for response $y_i~(i=1,\cdots,K)$, respectively. The token-level advantage for $y_i$ is formulated as follows:
\begin{equation}
A_{i,t}=\underbrace{r^v_i-\frac{1}{K-1}\sum_{k\neq i}r^v_k}_\text{RLOO with verifiable reward}+\underbrace{\sum_{j=t}^{N}{\gamma^{j-t} \cdot r^{p*}_{i,j}} -\underbrace{\frac{\sum_{k\neq i}\sum_{l=1}^N\sum_{j=l}^N {\gamma^{j-l} \cdot r^{p*}_{k,j}}}{(K-1)N}}_\text{token-level baseline} }_\text{RLOO with token-level transformed process rewards}
\label{eq:good adv}
\end{equation}
where $t=1,\cdots,N$. Specifically: \textbf{(i)} for verifiable rewards, we directly adopt RLOO. \textbf{(ii)} For process rewards, we employ a token-level baseline to avoid reward hacking, as discussed in the second case in \S~\ref{sec:reward hacking}. Moreover, we normalize the baseline with the max generation length $N$ instead of response length to avoid length biases, similar to concurrent work~\citep{liu2025understanding}.
% \vspace{-5pt}
\section{Experiments} \label{sec:experiment}
% \vspace{-2pt}
\subsection{PRM Training and Evaluation}
% \vspace{-3pt}
To conduct our RFT experiments, we first require a PRM to provide process rewards. We train our PRM based on Qwen2.5-Math-7B due to its strong performance in mathematical tasks. Following~\citet{lightman2023let}, we treat the PRM training as a binary classification task. We replace the final layer of the model with a value head and train the model on the PRM800K dataset~\citep{lightman2023let} in 2 stages. In the first stage, we freeze the LLM parameters and train only the value head with a learning rate $10^{-4}$ for 3 epochs. In the second stage, we unfreeze the LLM parameters and fine-tune all parameters with a learning rate $10^{-6}$ for 1 epoch.

\begin{wraptable}{r}{0.48\textwidth}
    \centering
    \small
    \vspace{-10pt}
    \captionof{table}{Results of BoN evaluation. Rows marked with * are taken from \citet{xiong2024rlhflowmath}.}
    \setlength{\tabcolsep}{2pt}{
    \begin{tabular}{l|cc}
    \specialrule{.1em}{.05em}{.05em}
        \textbf{Method} & \textbf{GSM8K} & \textbf{MATH} \\ \hline
        Pass@1 * & 83.9 & 42.4 \\
        Majority Voting@1024 * & 89.7 & 57.4 \\
        Deepseek-PRM-7B BoN@1024 * & \textbf{93.0} & 58.1 \\
        \rowcolor{lightblue}PURE-PRM-7B BoN@1024 & 91.6 & \textbf{62.6} \\ 
    \specialrule{.1em}{.05em}{.05em}
    \end{tabular}}
    \vspace{-10pt}
\end{wraptable}
We evaluate our PRM, named PURE-PRM-7B, through 3 ways: Best-of-N (BoN) method, ProcessBench~\citep{zheng2024processbench}, and PRMBench~\citep{song2025prmbench}. For BoN evaluation, we use rollout data from~\citet{xiong2024rlhflowmath}. For each question in GSM8K~\citep{cobbe2021training} and MATH~\citep{hendrycks2021measuring}, they uses Deepseek-7B to generate $K=1024$ answers. We score each step using our PRM, transform process rewards using Eq. (\ref{eq:transform reward}), and compute an outcome score by summing the transformed rewards for each answer. The answer with the highest outcome score is selected as the final answer. Our PURE-PRM-7B achieves BoN@1024 accuracy of 91.6\% on GSM8K and 62.6\% on MATH, performing comparably to the best results of 93.0\% and 58.1\% reported by~\citet{xiong2024rlhflowmath}

On ProcessBench~\citep{zheng2024processbench}, which assesses the PRM's ability to identify the first process error, our PRM achieves a state-of-the-art average F1 score of 57.5, surpassing the previous best F1 score of 56.5 reported in the benchmark. Detailed scores for each subset of ProcessBench are provided in Appendix \ref{app:eval_of_prm}. On PRMBench~\citep{song2025prmbench}, which evaluates the fine-grained error detection capabilities of PRMs, our PURE-PRM-7B ranks third among open-source PRMs\footnote{The official leaderboard is at \url{https://prmbench.github.io/}.}. These results demonstrate that our PRM achieves top performance and is suitable for fine-tuning LLMs.

% \vspace{-5pt}
\subsection{RL Settings}
\textbf{Reward types.}
Our framework supports 3 types of rewards: process reward only (PURE-PRM), verifiable reward only (PURE-VR), which matches the training setup of Deepseek R1-Zero, and a mix of both (PURE-PRM+VR). For verifiable rewards, a reward of +1 is assigned if the generated answer matches the ground-truth answer; otherwise, a reward of 0 is given. No format-related rewards.

\textbf{RL dataset.}
We use the RFT dataset from SimpleRL~\citep{zeng2025simplerl}. It samples around 8,000 problems from MATH dataset~\citep{hendrycks2021measuring} with difficulty lv.3-5. For ground-truth (GT) answers, PURE-PRM method does not use GT answers. PURE-VR method requires GT answers for all problems. For PURE-PRM+VR, GT answers are randomly assigned to 1/10 of the problems, leading to about 800 problems with GT answers and 7,200 open problems. This setup aims to explore the effectiveness of process rewards as the main signal for RL training.

\textbf{Hyperparameters.}
We use veRL~\citep{sheng2024hybridflow} to conduct experiments on 3 models: Qwen2.5-7B~\citep{yang2024qwen2}, Qwen2.5-Math-7B, Qwen2.5-Math-1.5B~\citep{yang2024qwen25math}. We use a constant learning rate of $10^{-6}$ for PURE-VR and PURE-PRM+VR and $5\times10^{-7}$ for PURE-PRM. Training steps are set to 500 for Qwen2.5-Math series and 1000 for Qwen2.5-7B. Other shared hyperparameters are set as follows: prompt batch size of 64, group size of 8 (generating 8 responses per prompt), training mini-batch size of 512, maximum generation length of 8192 tokens, sampling temperature of 1.0, KL coefficient of $10^{-3}$, transform temperature in Eq~\eqref{eq:transform reward} of $0.1$, save interval for checkpoints of 50 steps. 

\textbf{Baselines.} We compare our method with 3 state-of-the-art RFT methods: \textbf{(i)} Eurus-2-7B-PRIME~\citep{cui2025process}: A model based on Qwen2.5-Math-7B fine-tuned with implicit PRM. \textbf{(ii)} SimpleRL-Zoo~\citep{zeng2025simplerl}: A re-implementation of Deepseek R1's training recipe on several base models. \textbf{(iii)} Qwen2.5-7B-DPO-Zero~\citep{xiong2025self}: A model based on Qwen2.5-Math-7B fine-tuned with iterative DPO.

\textbf{Evaluation metrics.}
At test time, we evaluate performance on 5 competition-level mathematical benchmarks, including AIME24~\citep{li2024numinamath}, AMC23~\citep{li2024numinamath}, MATH500~\citep{hendrycks2021measuring}, Minerva Math~\citep{lewkowycz2022solving}, OlympiadBench~\citep{he2024olympiadbench}. We report scores of best checkpoint saved in training. During training, we track about 20 metrics, including accuracy, reward, response length, repetition score, etc. Details are provided in Appendix \ref{app:details_of_metrics}.

\begin{table}[t]
\centering
\small
\captionof{table}{Detailed performance of various models across 5 benchmarks. Report pass@1 accuracy tested with greedy decoding. The blue lines represent the models trained with our recipe.}
\vspace{5pt}
\resizebox{\textwidth}{!}{
\begin{tabular}{l|l|ccccc|c}
\toprule
\multirow{2}{*}{\textbf{Base Model}} & \multirow{2}{*}{\textbf{Method}} & \textbf{MATH} & \textbf{Minerva} & \textbf{Olympiad} & \multirow{2}{*}{\textbf{AIME24}} & \multirow{2}{*}{\textbf{AMC23}} & \multirow{2}{*}{\textbf{Avg.}} \\
 & & \textbf{500} & \textbf{Math} & \textbf{Bench} & & & \\
\midrule
\multirow{5}{*}{Qwen2.5-7B} & SimpleRL-Zoo & 78.4 & 31.2 & 39.1 & 16.7 & 50.0 & 43.1 \\
 & Base & 71.4 & 23.9 & 35.3 & 10.0 & 52.5 & 38.6 \\
 & \cellcolor{lightblue}+ PURE-PRM & \cellcolor{lightblue}76.2 & \cellcolor{lightblue}37.1 & \cellcolor{lightblue}41.2 & \cellcolor{lightblue}13.3 & \cellcolor{lightblue}60.0 & \cellcolor{lightblue}45.6 \\
 & \cellcolor{lightblue}+ PURE-VR & \cellcolor{lightblue}75.6 & \cellcolor{lightblue}32.7 & \cellcolor{lightblue}39.0 & \cellcolor{lightblue}16.7 & \cellcolor{lightblue}55.0 & \cellcolor{lightblue}43.8 \\
 & \cellcolor{lightblue}+ PURE-PRM+VR & \cellcolor{lightblue}80.4 & \cellcolor{lightblue}37.9 & \cellcolor{lightblue}43.0 & \cellcolor{lightblue}16.7 & \cellcolor{lightblue}60.0 & \cellcolor{lightblue}\textbf{47.6} \\
\hline
% ---
\multirow{7}{*}{Qwen2.5-Math-7B} & Eurus-2-7B-PRIME & 79.2 & 38.6 & 42.1 & 26.7 & 57.8 & 48.9 \\
 & SimpleRL-Zoo & 80.2 & 38.2 & 43.3 & 23.3 & 55.0 & 48.0 \\
 & Qwen2.5-7B-DPO-Zero & 76.8 & 30.9 & 37.9 & 26.7 & 62.5 & 47.0 \\
 & Base & 71.8 & 29.8 & 35.1 & 13.3 & 47.5 & 39.5 \\
 & \cellcolor{lightblue}+ PURE-PRM & \cellcolor{lightblue}81.8 & \cellcolor{lightblue}38.2 & \cellcolor{lightblue}44.7 & \cellcolor{lightblue}16.7 & \cellcolor{lightblue}60.0 & \cellcolor{lightblue}49.3 \\
 & \cellcolor{lightblue}+ PURE-VR & \cellcolor{lightblue}79.4 & \cellcolor{lightblue}36.8 & \cellcolor{lightblue}41.8 & \cellcolor{lightblue}23.3 & \cellcolor{lightblue}60.0 & \cellcolor{lightblue}48.3 \\
 & \cellcolor{lightblue}+ PURE-PRM+VR & \cellcolor{lightblue}82.6 & \cellcolor{lightblue}37.1 & \cellcolor{lightblue}44.1 & \cellcolor{lightblue}20.0 & \cellcolor{lightblue}82.5 & \cellcolor{lightblue}\textbf{53.3} \\
\hline
% ---
\multirow{4}{*}{Qwen2.5-Math-1.5B} & Base & 61.4 & 23.5 & 29.3 & 13.3 & 40.0 & 33.5 \\
 & \cellcolor{lightblue}+ PURE-PRM & \cellcolor{lightblue}75.2 & \cellcolor{lightblue}26.5 & \cellcolor{lightblue}36.4 & \cellcolor{lightblue}13.3 & \cellcolor{lightblue}50.0 & \cellcolor{lightblue}40.3 \\
 & \cellcolor{lightblue}+ PURE-VR & \cellcolor{lightblue}74.2 & \cellcolor{lightblue}27.9 & \cellcolor{lightblue}36.0 & \cellcolor{lightblue}10.0 & \cellcolor{lightblue}55.0 & \cellcolor{lightblue}40.6 \\
 & \cellcolor{lightblue}+ PURE-PRM+VR & \cellcolor{lightblue}76.0 & \cellcolor{lightblue}31.6 & \cellcolor{lightblue}37.2 & \cellcolor{lightblue}16.7 & \cellcolor{lightblue}55.0 & \cellcolor{lightblue}\textbf{43.3} \\
\bottomrule
\end{tabular}
}
\label{tab:main results}
\vspace{-10pt}
\end{table}
\vspace{-5pt}
\subsection{Main Results} \label{sec:main results}
We report the pass@1 accuracy across 5 benchmarks in Table \ref{tab:main results}. The results indicate that all variants of our methods perform at least comparable to, or better than baselines. For example, when using only verifiable rewards, following the setup of SimpleRL-Zoo, our method PURE-VR achieves average scores of 43.8 and 48.3 on Qwen2.5-7B and Qwen2.5-Math-7B, respectively. These scores are comparable to the baseline results of 43.1 and 48.0 obtained by SimpleRL-Zoo, confirming the reliability of our code-base. Based on Table \ref{tab:main results}, we draw the following observation:

\textbf{PRM-based approach performs similar to the VR-based approach, but combining the two yields better results.}
On the Qwen2.5-Math-7B model, PURE-PRM achieves an average score of 49.3, slightly higher than the 48.3 of PURE-VR. However, the combined method, PURE-PRM+VR, reaches a score of 53.3, surpassing PURE-VR by approximately 5 percentage points. This trend remains consistent across the other 2 base models.

Next, we analyze the training dynamics of our methods to understand the limitations of the PRM-based approach and explain why PURE-PRM+VR is superior. Figure \ref{fig:main results training curves} illustrates the training curves for five variants of our methods on Qwen2.5-Math series, leading to the following findings:

\begin{figure}[htb]
    \centering
    \begin{minipage}{0.66\textwidth}
        \centering
        \includegraphics[width=\linewidth]{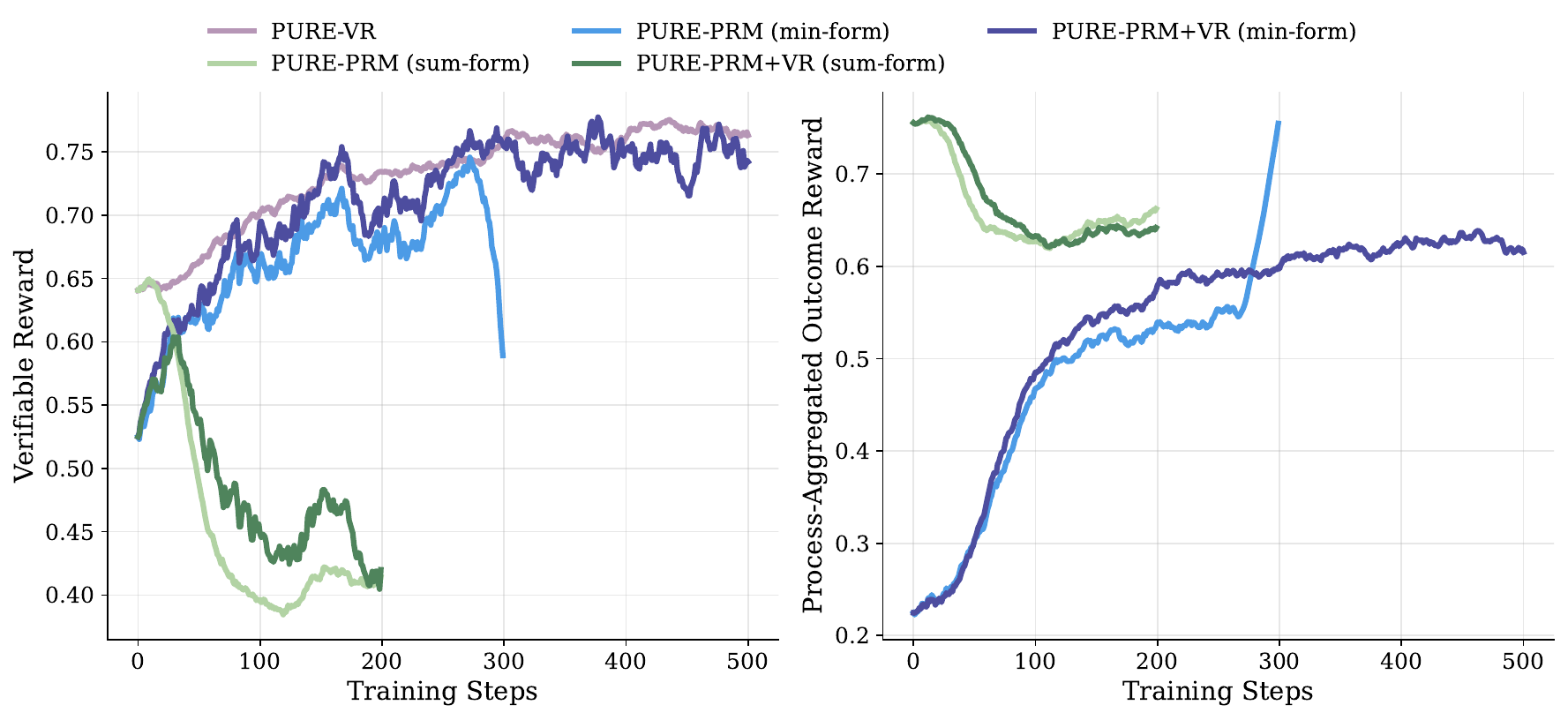}
        (a) PURE on Qwen2.5-Math-7B
    \end{minipage}
    \hfill
    \begin{minipage}{0.32\textwidth}
        \centering
        \includegraphics[width=\linewidth]{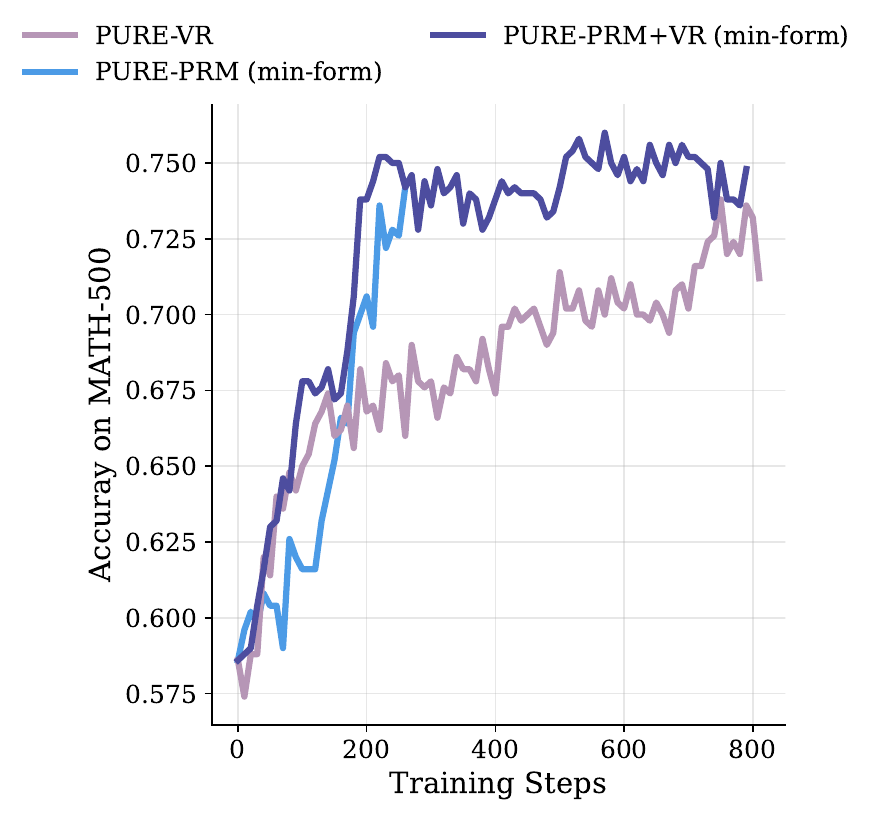}
        (b) Qwen2.5-Math-1.5B
    \end{minipage}
    \caption{Training curves for different variants of our methods on Qwen2.5-Math series. Curves of PURE-PRM (sum-form) and PURE-PRM+VR (sum-form) are truncated due to training collapse. Process-aggregated outcome reward is the summation of final process rewards for one response: for sum-form, it sums PRM-emitted rewards; for min-form, it sums transformed rewards, approximating the minimum PRM-emitted reward. Thus \textbf{values across the 2 credit assignment methods are not comparable}. For PURE-PRM, verifiable reward is logged but unused in training.}
    \label{fig:main results training curves}
    \vspace{-7pt}
\end{figure}

\textbf{Summation-form credit assignment collapses training even at the beginning, while the min-form method significantly enhances training stability.}
As shown in Figure \ref{fig:main results training curves}a, both PURE-PRM (sum-form) and PURE-PRM+VR (sum-form) experience collapse at step 25. At step 80, their average benchmark scores drop to around 30, which is much lower than the base model’s score of 39.5. In contrast, the min-form methods remain stable over 200 steps, achieving average scores of 49.3 and 53.3 for PURE-PRM (min-form) and PURE-PRM+VR (min-form), respectively.

\textbf{Dense rewards substantially improve learning efficiency compared to sparse rewards.} Figure \ref{fig:main results training curves}b shows the curves of accuracy on MATH-500 for 3 variants of our methods. We find that PRM-involved approaches takes around 30\% of the training steps to achieve the same accuracy as PURE-VR.

\textbf{Reward hacking is inevitable when rely solely on PRM, though it can be delayed through algorithmic adjustments.}
While PURE-PRM (min-form) produces a well-tuned model, reward hacking still occurs at step 270, where verifiable rewards decrease sharply while process-aggregated outcome rewards increase in Figure \ref{fig:main results training curves}a. This issue was observed in all experiments using the PURE-PRM method, indicating that reward hacking is a persistent challenge when relying solely on PRM and limits further progress. However, compared to sum-form credit assignment, min-form approaches delays the onset of reward hacking significantly. Such reward hacking corresponds to the third case in \S~\ref{sec:reward hacking}, which is inevitable due to the current architecture of PRMs.

\textbf{Incorporating a few ground-truth signals can effectively reduce reward hacking.}
PURE-PRM+VR (min-form) uses a mixed dataset of 800 problems with GT answers and 7,200 open problems. Thus, during training, process rewards guide the RFT process primarily, while verifiable rewards serve as an auxiliary signal. As shown in Figure \ref{fig:main results training curves}a, this approach achieves a stable training process similar to PURE-VR and attains the highest average benchmark score of 53.3. This suggests a practical solution to address reward hacking by including a small proportion (around 10\%) of ground-truth signals during the RFT stage.
\vspace{-5pt}
\section{Analysis} \label{sec:analysis}
\vspace{-3pt}
\subsection{Reward Hacking induced by PRM} \label{sec:reward hacking}
\vspace{-2pt}
In this section, we detail the reward hacking cases observed during training. We categorize these cases into the 3 types. Specific examples for each type are provided in Appendix \ref{app:reward_hacking_examples}.

\textbf{Case 1: Only thinking, not solving.}
This behavior occurs when steps with certain patterns, such as thinking, are rewarded significantly more than others, and the inappropriate credit assignment further widen the gap in advantages between steps. As a result, the model learns to exploits these patterns to achieve higher scores. As shown in Figure \ref{fig:difference of credit assignment}, the sum-form credit assignment increases the likelihood of generating thinking steps while reducing that of incorrect solution steps, resulting in an armchair general. However, exploiting specific patterns is not always negative. For example, using patterns like backtracking and verification improves reasoning skills~\citep{guo2025deepseek,gandhi2025cognitive}.

\textbf{Case 2: Extremely few steps (1 step).}
This happens when an unsuitable advantage baseline is used. In \S~\ref{sec:advantage estimation}, we employ a token-level baseline for process rewards. An alternative approach is a step-level baseline, formulated as follows:
\begin{equation}
    A_{i,t}=r^v_i-\frac{1}{K-1}\sum_{k\neq i}r^v_k+\sum_{j=t}^{N} \gamma^{j-t} \left[ r^{p*}_{i,j} - \frac{1}{K-1}\sum_{k\neq i} \frac{\sum_l r^{p*}_{k,l}}{|y_k|_s} \right]
    \label{eq:bad adv baseline}
\end{equation}
where $|y_k|_s$ represents the number of steps of the $k$-th response within the group. The advantage baseline for process rewards in Eq. (\ref{eq:bad adv baseline}) means the average process reward of other responses in the group. However, we find the step-level baseline is biased against the number of steps. When the process-aggregated outcome reward are equal, response with more steps are penalized more heavily by the baseline, causing the model to favor responses with fewer steps. Eventually, the model learns to output only a single step with an excessively large number of tokens. This undermines the purpose of PRM, which is to evaluate the process step by step.

\textbf{Case 3: Extremely few steps (0 step).}
In this case, the model learns to output irrelevant responses such as ``Thank you.’’, ``Happy Birthday.’’, or even empty responses. Since PRM infers in a causal manner, it assigns high rewards to these meaningless steps, not realizing that no further content follows. This issue stems from the current architecture of discriminative PRMs (\textit{i.e.}, the causal attention mask). Our solution is adding GT-level signal as an aid, such as verifiable rewards in PURE-PRM+VR. Another potential solution is the generative PRMs, which we leave for future work.

\begin{figure}[th]
    \centering
    \includegraphics[width=0.9\textwidth]{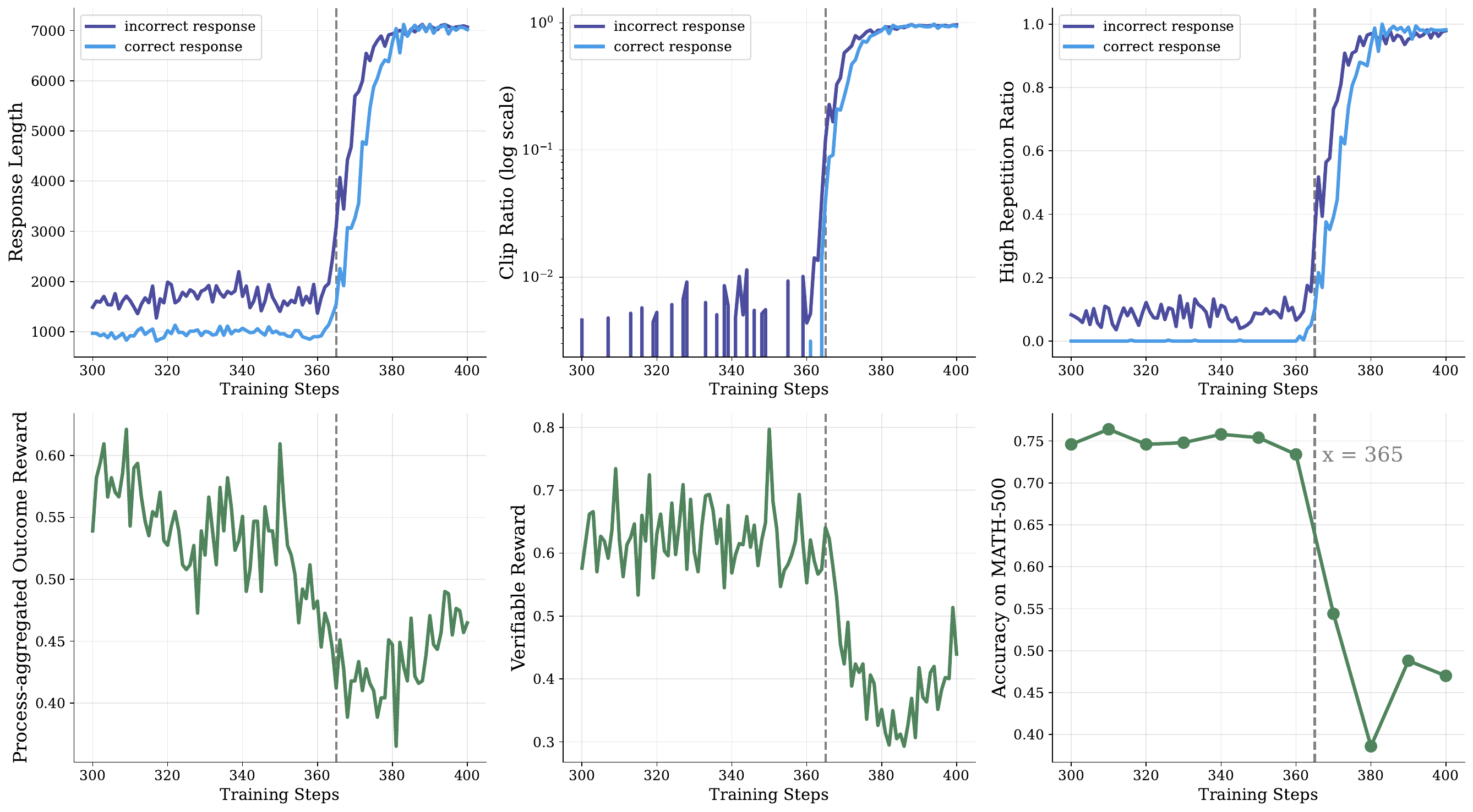}
    \caption{Training curves for PURE-PRM+VR with doubled process rewards based on Qwen2.5-7B. The correctness of responses are judged by the verifier. Process-aggregated outcome reward (bottom-left) is the summation of transformed process rewards for one response. No smooth is applied. Training collapses at step 365, showing a sharp drop of rewards and accuracy.}
    \label{fig:correlation_high_level_metrics}
    \vspace{-10pt}
\end{figure}
% \vspace{-3pt}
\subsection{Causes of Training Collapse} \label{sec:cause of training collapse}
% \vspace{-2pt}
In this section, we train Qwen2.5-7B using PURE-PRM+VR with doubled process rewards, and show the training curves of pattern-related metrics in Figure \ref{fig:correlation_high_level_metrics}, including response length, clip ratio, repetition score, rewards, and accuracy on MATH-500. The clip ratio measures how often responses are cut off due to the maximum generation length limit. The repetition score evaluates how repetitive a response is by calculating the longest common prefix (LCP) lengths between all pairs of its suffixes~\footnote{This is implemented using a function from Open-Reasoner-Zero~\citep{hu2025openreasonerzeroopensourceapproach}.}. We consider responses with repetition score above 0.2 as highly repetitive. Details on metrics are provided in Appendix \ref{app:details_of_metrics}.

As shown in Figure \ref{fig:correlation_high_level_metrics}, training collapses at step 365, showing a sharp drop of rewards and accuracy. Before step 360, incorrect responses judged by the verifier are longer and more repetitive than the correct responses, but that does not collapse the training. Both the clip ratio and high repetition ratio for correct responses remain at 0 until step 361. After that, the response length, clip ratio, and high repetition ratio for both incorrect and correct responses rise sharply until step 380. Based on these observations, we derive the following conclusions:

\textbf{Long and highly repetitive ``correct’’ responses cause training collapse.}
At step 361, the clip ratio and high repetition ratio for correct responses become greater than 0. This large number of positive signals for repetitive patterns is fatal to training. We refer to these samples as pseudo-positive samples. We attempt to treat pseudo-positive samples as incorrect samples and assign 0 rewards instead of +1 rewards, but this does not help much. The model learns to repeat content in ways that the LCP function can not detect, as detailed in Appendix \ref{app:details_of_metrics}. Verifiers can not identify pseudo-positive samples. While PRM has the potential to detect them, it currently fails to do so because such patterns are not included in its training data. This highlights a need for future PRM development to not only assess the correctness of steps but also evaluate the quality of patterns. 

\textbf{The model learns repetitive patterns quickly, leading to collapse within 5 gradient steps.}
From the unusual metrics at step 361 to the training collapse at step 365, the model rapidly learns to repeat content. Our training setup ensures one gradient step per training step, meaning the collapse happens within just 5 gradient steps.

% \subsection{Online Curriculum Learning vs. Offline model-based data filtering}

% \subsection{Ablation on Process Temperature}

% \subsection{Ablation on Data Amount}

% \vspace{-5pt}
\section{Conclusion} \label{sec:conclusion}
% \vspace{-5pt}
In this paper, we present PURE: \underline{\textbf{P}}rocess s\underline{\textbf{U}}pervised \underline{\textbf{R}}einforcement l\underline{\textbf{E}}arning, a framework leverages process rewards to improve the reasoning abilities of LLMs. Extensive experiments demonstrate that PURE effectively alleviates reward hacking induced by PRM, attributed to the proposed min-form credit assignment. This method allows PRM-based RFT to achieve performance similar to verifiable rewards-based approaches within 30\% steps, and further outperform them with the assistance of a few ground-truth signals. Additionally, we summarize the observed reward hacking cases during training, and find that pseudo-positive samples collapse training.

There are several promising directions for further improving PRM-based RFT. First, developing generative PRMs is both urgent and crucial. As discussed in \S~\ref{sec:analysis}, current PRMs are unable to address the third type of reward hacking mentioned in \S~\ref{sec:reward hacking}, and also cannot evaluate the quality of patterns like endless repetition. Generative PRMs, however, could potentially resolve these issues by making better use of the strong language capabilities of LLMs. Second, iterative training between PRM and LLM is essential to ensure that the PRM continuously adapts to the output distribution of LLMs.

\section*{Acknowledgments}
We sincerely thank Ganqu Cui, Songjun Tu, and Zhengbo Wang for their suggestions to the early draft of the paper. This work was partially supported by the National Science and Technology Major Project (2022ZD0117102), the National Natural Science Foundation of China under Grants 62271485 and 62303462, Beijing Natural Science Foundation under grant L241016 and L233005, Chongqing Transportation Technology Project (CQJT-CZKJ2024-04), the Provincial Key Research and Development Program of Zhejiang (project number: 2022C01129), Ningbo International Science and Technology Cooperation Project (2023H020), CCF-Tencent Rhino-Bird Open Research Fund.

\bibliography{neurips_2025}
\bibliographystyle{neurips_2025}

\appendix

\section{Related Work}
\subsection{Reinforcement Fine-Tuning}
Reinforcement fine-tuning (RFT) is a promising technique to improve the performance of LLMs~\citep{ouyang2022training, yue2024sc,wei2025swe}. OpenAI’s o1 model~\citep{jaech2024openai} was among the first to demonstrate the significant potential of large-scale reinforcement learning for enhancing reasoning abilities in LLMs. Recent studies have further confirmed that a straightforward reinforcement learning approach, using verifiable rewards, can scale effectively~\citep{guo2025deepseek, team2025kimi, zeng2025simplerl}. However, previous research has faced difficulties in effectively utilizing PRM~\citep{guo2025deepseek} in training-time, which is the primary focus of our work in PURE.

% \vspace{-5pt}
\subsection{Reward Models for LLM Alignment}
In the area of LLM alignment, outcome reward models (ORMs) are used to evaluate the quality of the entire response generated by LLMs. Employing ORMs for aligning LLMs with human values has become a common practice in the post-training stage~\citep{ouyang2022training, bai2023qwen, touvron2023llama}. However, ORMs fall short in providing detailed feedback for complex tasks that involve multiple reasoning steps. In contrast, process reward models (PRMs) provide more detailed feedback by evaluating each step of the reasoning process, allowing LLMs to learn more effectively from mistakes made during reasoning~\citep{lightman2023let, yuan2024free, cui2025process}. Studies have shown that PRMs outperform majority voting and ORMs in test-time scaling~\citep{lightman2023let, wang2023math, xiong2024rlhflowmath}. Nevertheless, the application of PRMs during the training phase remains largely unexplored, which is the central topic of PURE.

\section{Proof of Q-value estimation error} \label{app:proof}
\begin{theorem}[\textbf{Q-Value Estimation Error Bound Comparison}]
Under Assumptions 1 and 2, for any state-action pair $(s_t, a_t)$ and a trajectory $\tau$ with $n$ reasoning steps:
\begin{enumerate}[leftmargin=*]
    \item \textbf{Sum-form Error Bound:} The estimation error for the summation-form Q-value, is bounded by:
    $$
    |Q_{\pi}^{sum}(s_t, a_t) - Q_{\pi}^{sum,*}(s_t, a_t)| \le \sum_{i=t}^{t+n-1} \gamma^{i-t} \epsilon
    $$
    For an infinite horizon ($n \to \infty$), this bound becomes:
    $$
    |Q_{\pi}^{sum}(s_t, a_t) - Q_{\pi}^{sum,*}(s_t, a_t)| \le \frac{\epsilon}{1 - \gamma}
    $$

    \item \textbf{Min-form Error Bound:} The estimation error for the min-form Q-value, derived from Eq. (\ref{eq:min-form Q}), is bounded by:
    $$
    |Q_{\pi}^{min}(s_t, a_t) - Q_{\pi}^{min,*}(s_t, a_t)| \le \epsilon
    $$
\end{enumerate}
\end{theorem}

\begin{proof}[\textbf{Proof}]

\textbf{Part 1: Proof for the Sum-form Error Bound.}

We begin with the definition of the absolute error of the Q-value, using the canonical formulation in Eq. (\ref{eq:definition of classical Q}). $Q_{\pi}^{sum}$ denotes the value calculated with the estimated rewards from the PRM ($r^p$), and $Q_{\pi}^{sum,*}$ denotes the true value (calculated with $r^*$).
    
\begin{align*}
    |Q_{\pi}^{sum}(s_t, a_t) - Q_{\pi}^{sum,*}(s_t, a_t)| &= |E_{\pi \sim \tau}[\sum_{i=t}^{t+n-1} \gamma^{i-t} r_i^p] - E_{\pi \sim \tau}[\sum_{i=t}^{t+n-1} \gamma^{i-t} r_i^*]| \\
    &= |E_{\pi \sim \tau}[\sum_{i=t}^{t+n-1} \gamma^{i-t} (r_i^p - r_i^*)]| \\
    &\le E_{\pi \sim \tau}[|\sum_{i=t}^{t+n-1} \gamma^{i-t} (r_i^p - r_i^*)|] \\
    &\le E_{\pi \sim \tau}[\sum_{i=t}^{t+n-1} |\gamma^{i-t} (r_i^p - r_i^*)|] \\
    &= E_{\pi \sim \tau}[\sum_{i=t}^{t+n-1} \gamma^{i-t} |r_i^p - r_i^*|]
\end{align*}

Now, we apply Assumption 1, which states that the reward error at each step is bounded by $\epsilon$, i.e., $|r_t^p - r_t^*| \le \epsilon$.
\begin{equation*}
    |Q_{\pi}^{sum}(s_t, a_t) - Q_{\pi}^{sum,*}(s_t, a_t)| \le E_{\pi \sim \tau}[\sum_{i=t}^{t+n-1} \gamma^{i-t} \epsilon] = \epsilon \sum_{i=t}^{t+n-1} \gamma^{i-t}
\end{equation*}
For the infinite horizon case, the geometric series converges to $\frac{1}{1-\gamma}$.

\textbf{Part 2: Proof for the Min-form Error Bound.}

We start similarly with the definition of the absolute Q-value error for the min-form. The return $G(s_t, a_t|\tau)$ is the value being expected.
\begin{align*}
    |Q_{\pi}^{min}(s_t, a_t) - Q_{\pi}^{min,*}(s_t, a_t)| &= |E_{\pi \sim \tau}[G(s_t, a_t|\tau)] - E_{\pi \sim \tau}[G^*(s_t, a_t|\tau)]| \\
    &= |E_{\pi \sim \tau}[G(s_t, a_t|\tau) - G^*(s_t, a_t|\tau)]|
\end{align*}
    
Now we analyze the return term inside the expectation. According to Equation (3), for a trajectory $\tau$ and a step $t$, the return is defined as the minimum of future rewards.
\begin{align*}
    G(s_t, a_t|\tau) &= \min(r_t^p, r_{t+1}^p, \ldots, r_n^p) \\
    G^*(s_t, a_t|\tau) &= \min(r_t^*, r_{t+1}^*, \ldots, r_n^*)
\end{align*}
    
We use a key property of the minimum function: $|\min(a) - \min(b)| \le \max_i |a_i - b_i|$. Applying this property:
\begin{equation*}
    |G(s_t, a_t|\tau) - G^*(s_t, a_t|\tau)| = |\min_{i=t \ldots n}(r_i^p) - \min_{i=t \ldots n}(r_i^*)| \le \max_{i=t \ldots n} |r_i^p - r_i^*| \le \epsilon
\end{equation*}
    
Substitute this result back into the expectation inequality:
\begin{equation*}
    |Q_{\pi}^{min}(s_t, a_t) - Q_{\pi}^{min,*}(s_t, a_t)| = |E_{\pi \sim \tau}[G(s_t, a_t|\tau) - G^*(s_t, a_t|\tau)]| \le E_{\pi \sim \tau}[\epsilon] = \epsilon
\end{equation*}
This completes the proof. It is proven that the error in the min-form does not accumulate and is directly bounded by the maximum single-step error of the reward model.
\end{proof}

\section{Benchmark Scores of Our PRM} \label{app:eval_of_prm}
We report the detailed scores on ProcessBench~\citep{zheng2024processbench} and PRMBench~\citep{song2025prmbench} in Table \ref{tab:process_bench_results} and Table \ref{tab:prmbench_results}, respectively.
\begin{table}[!ht]
  \centering
  \caption{
  F1 scores of each subset in ProcessBench. The blue line represents our PRM. Other lines are taken from \citet{zheng2024processbench}.
  }
    \begin{tabular}{lccccc}
    \toprule
    \multirow{2}{*}{\textbf{Model}} & \multirow{2}{*}{\textbf{GSM8K}} & \multirow{2}{*}{\textbf{MATH}} & \textbf{Olympiad-} & \textbf{Omni-} & \multirow{2}{*}{\textbf{Average}} \\
     & & & \textbf{Bench} & \textbf{MATH} & \\
    \midrule
    Math-Shepherd-PRM-7B & 47.9  & 29.5  & 24.8  & 23.8  & 31.5  \\
    RLHFlow-PRM-Deepseek-8B & 38.8  & 33.8  & 16.9  & 16.9  & 26.6  \\
    Skywork-PRM-7B & \textbf{70.8} & 53.6  & 22.9  & 21.0  & 42.1  \\
    Qwen2.5-Math-7B-PRM800K & 68.2  & 62.6 & \textbf{50.7} & 44.3 & 56.5 \\
    \rowcolor{lightblue}PURE-PRM-7B & 69.0 & \textbf{66.5} & 48.4 & \textbf{45.9} & \textbf{57.5} \\
    \bottomrule
    \end{tabular}
  \label{tab:process_bench_results}
\end{table}
\begin{table}[!ht]
  \centering
  \caption{
  Results on ProcessBench. The blue line represents our PRM. Other lines are taken from \citet{song2025prmbench}.
  }
    \begin{tabular}{lcccc}
    \toprule
    \textbf{Model} & \textbf{S1 (Simplicity)} & \textbf{S2 (Soundness)} & \textbf{S3 (Sensitivity)} & \textbf{Overall} \\
    \midrule
    MathShepherd-Mistral-7B & 47.1  & 45.7 & 60.7 & 47.0 \\
    RLHFlow-PRM-Deepseek-8B & 47.6 & 57.5 & 68.1 & 54.2 \\
    Skywork-PRM-7B & \textbf{59.6} & 68.5 & 73.3 & 65.1 \\
    Qwen2.5-Math-7B-PRM800K & 52.1  & \textbf{71.0} & 75.5 & \textbf{65.5} \\
    \rowcolor{lightblue}PURE-PRM-7B & 52.2 & 70.2 & \textbf{75.8} & 65.3 \\
    \bottomrule
    \end{tabular}
  \label{tab:prmbench_results}
\end{table}

\section{Details of Training Metrics} \label{app:details_of_metrics}
We detail 4 metrics used in the main text of the paper, including process-aggregated outcome reward, clip ratio, repetition score, and high repetition ratio.
\paragraph{Process-aggregated outcome reward.}
This metric reflects the value of a response with respect to process rewards. In practice, it is calculated as the sum of transformed process rewards for the min-form credit assignment, approximating the minimum PRM-emitted reward for a given response.

\paragraph{Clip ratio.}
The clip ratio indicates the proportion of samples in the replay buffer that are truncated due to the maximum generation length limit. After training collapses, LLMs often generate infinitely repetitive responses until hit the maximum context length.

\paragraph{Repetition score.}
The repetition score measures the level of repetition in a response by calculating the longest common prefix (LCP) lengths between all pairs of its suffixes. However, this method can only identify basic repetition patterns. A more complex form of repetition involves the use of near-synonyms, where words differ but the meaning remains the same. Such cases are not detected by the LCP method. This limitation explains why treating pseudo-positive samples as incorrect still results in training collapse, as mentioned in \S~\ref{sec:cause of training collapse}. The model learns to exploit complex repetition patterns to bypass the LCP function.

\paragraph{High repetition ratio.}
We consider responses with a repetition score greater than 0.2 as highly repetitive. This metric shows the percentage of samples in the replay buffer that fall into this category.

\section{Examples of Reward Hacking} \label{app:reward_hacking_examples}
\begin{figure}[htbp]
    \centering
    \includegraphics[width=\textwidth]{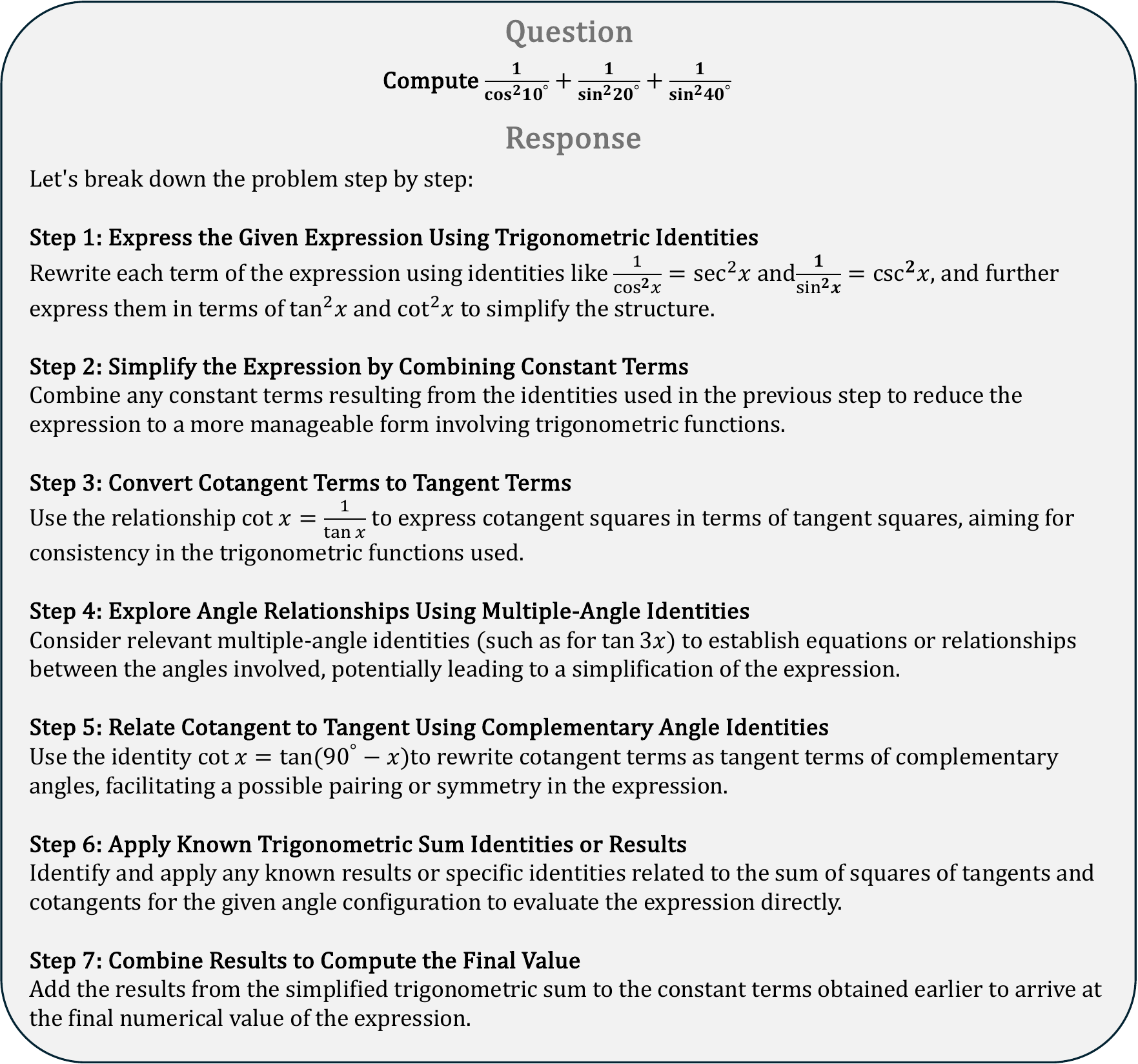}
    \caption{Reward hacking, case 1: only thinking, not solving. In this example, the LLM analyzes the problem and gives a few equations for trigonometric simplifications, but does not substitute actual numbers to calculate and solve the problem. This is because the LLM hacks the implicit pattern inside high-reward steps, \textit{i.e.}, thinking.}
    \label{fig:reward_hacking_case_1}
\end{figure}

\begin{figure}[htbp]
    \centering
    \includegraphics[width=\textwidth]{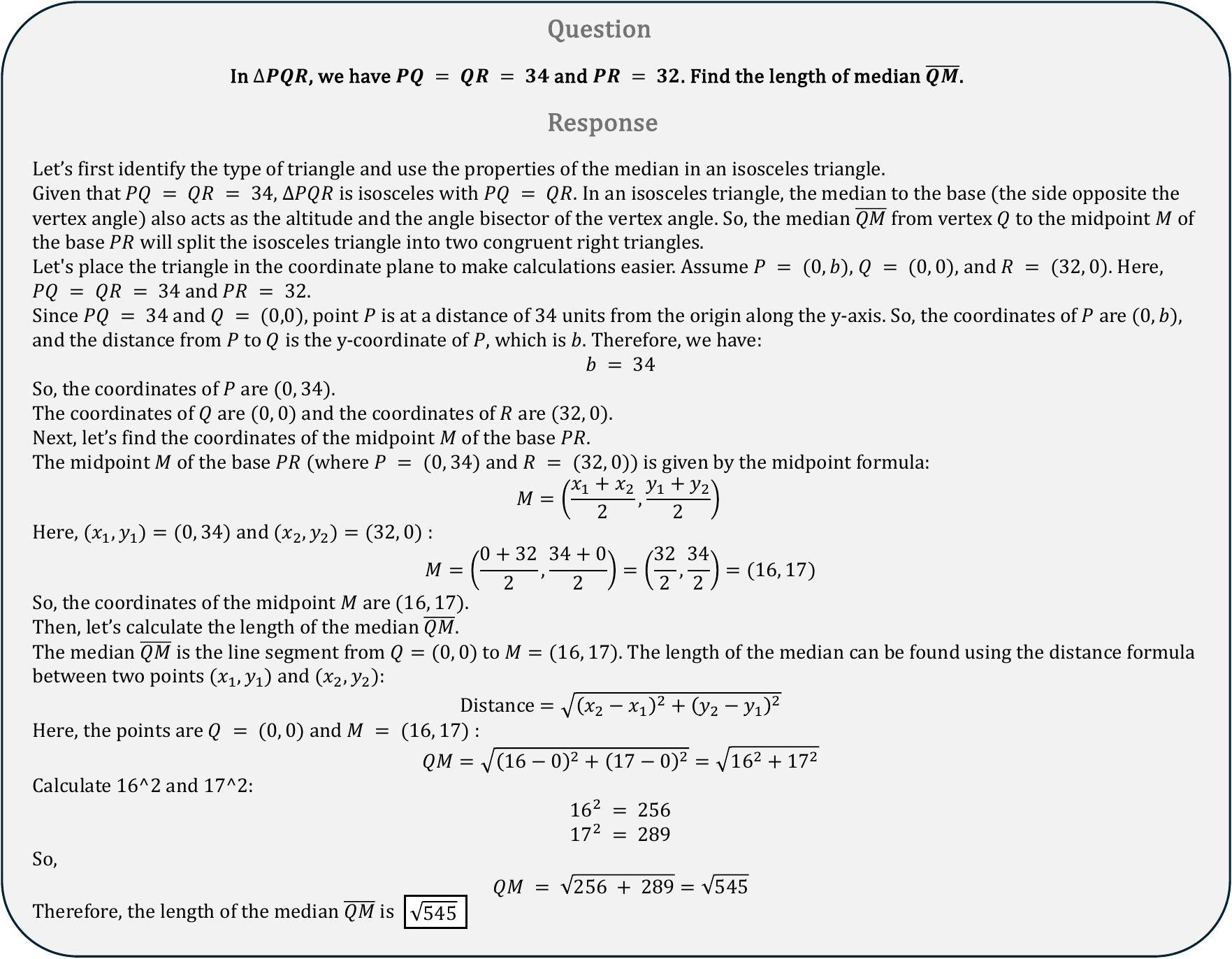}
    \caption{Reward hacking, case 2: extremely few steps (1 step). In practice, we divide steps according to double line breaks "$\backslash$n$\backslash$n" and then PRM scores each step. When the advantage baseline is inappropriate, such as step-level baseline discussed in \S~\ref{sec:reward hacking}, the model learns to deliberately avoid outputting "$\backslash$n$\backslash$n", preferring short-step response. In this example, there is no "$\backslash$n$\backslash$n" character in the generated response, resulting in the entire response being split into only one step.}
    \label{fig:reward_hacking_case_2}
\end{figure}

\begin{figure}[htbp]
    \centering
    \includegraphics[width=\textwidth]{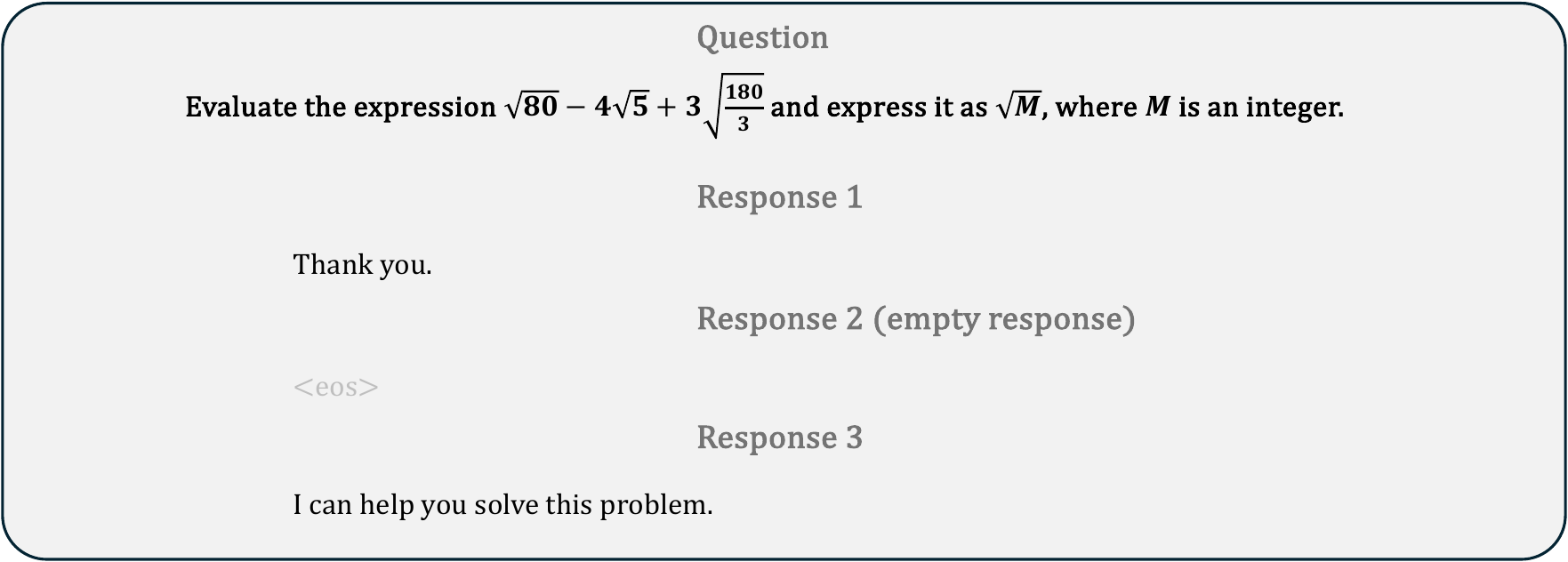}
    \caption{Reward hacking, case 3: extremely few steps (0 step). This is the most common cases for PURE-PRM. When relying solely on the PRM, training eventually boils down to this case after numerous steps of training. It is inevitable because the PRM scores based on the question and historical steps, and it does not know the role of the current step in the overall response.}
    \label{fig:reward_hacking_case_3}
\end{figure}

We show examples for each reward hacking case in Figure \ref{fig:reward_hacking_case_1}, \ref{fig:reward_hacking_case_2}, and \ref{fig:reward_hacking_case_3}. In practice, we use double line breaks ("$\backslash$n$\backslash$n") to separate individual steps within the response.

\section{Additional Experiments} \label{app:experiments}
\subsection{Ablation on Transform Temperature}

\begin{figure}
    \centering
    \includegraphics[width=0.4\linewidth]{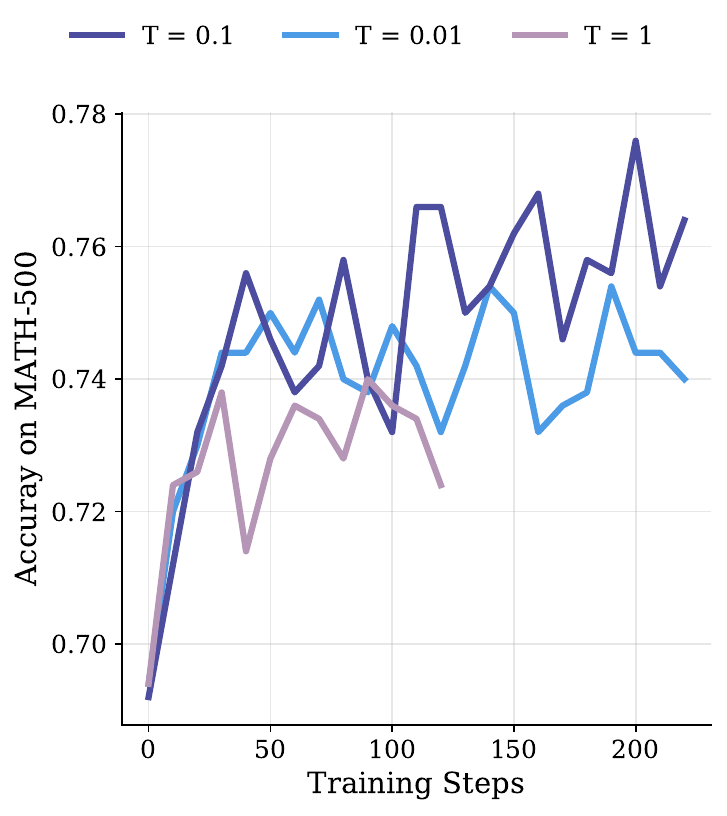}
    \caption{Ablation on transform temperature defined in Eq.~\eqref{eq:transform reward}. We use PURE-PRM+VR to fine-tune Qwen2.5-7B in this experiment.}
    \label{fig:ab_T}
\end{figure}
In this section, we ablate the transform temperature defined in Eq.~\eqref{eq:transform reward}, which controls the level of approximation to the min-form credit assignment. We choose Qwen2.5-7B as the base model and run PURE-PRM+VR with 3 values of transform temperature individually for 230 steps. The results are shown in Table \ref{tab:ab_T} and Figure \ref{fig:ab_T}, which indicates that 0.1 is the best for the transform temperature.
% \begin{figure}[htbp]
%     \centering
%     \includegraphics[width=\textwidth]{imgs/ab_transform_temperature.pdf}
%     \caption{Ablation on transform temperature defined in Eq.~\eqref{eq:transform reward}. We use PURE-PRM+VR to fine-tune Qwen2.5-7B in this experiment.}
%     \label{fig:ab_T}
% \end{figure}
\begin{table}[htbp]
\centering
\captionof{table}{Results of PURE-PRM+VR with different transform temperature, defined in Eq~\eqref{eq:transform reward}. We conduct the experiments based on Qwen2.5-7B and report pass@1 accuracy tested with greedy decoding.}
\vspace{5pt}
\small
\begin{tabular}{l|ccccc|c}
\toprule
\multirow{2}{*}{\textbf{Method}} & \textbf{MATH} & \textbf{Minerva} & \textbf{Olympiad} & \multirow{2}{*}{\textbf{AIME24}} & \multirow{2}{*}{\textbf{AMC23}} & \multirow{2}{*}{\textbf{Avg.}} \\
 & \textbf{500} & \textbf{Math} & \textbf{Bench} & & & \\
\midrule
  Base & 71.4 & 23.9 & 35.3 & 10.0 & 52.5 & 38.6 \\
 + PURE-PRM+VR $(T=0.01)$ & \textbf{76.2} & \textbf{37.1} & 39.0 & 6.7 & 55.0 & 42.8 \\
 + PURE-PRM+VR $(T=0.1)$ & \textbf{76.2} & \textbf{37.1} & \textbf{41.2} & \textbf{13.3} & \textbf{60.0} & \textbf{45.6} \\
 + PURE-PRM+VR $(T=1.0)$ & 75.8 & 33.5 & 38.7 & \textbf{13.3} & 52.5 & 42.8 \\
\bottomrule
\end{tabular}
\label{tab:ab_T}
\vspace{-5pt}
\end{table}

\begin{figure}[hb]
    \centering
    \includegraphics[width=0.8\textwidth]{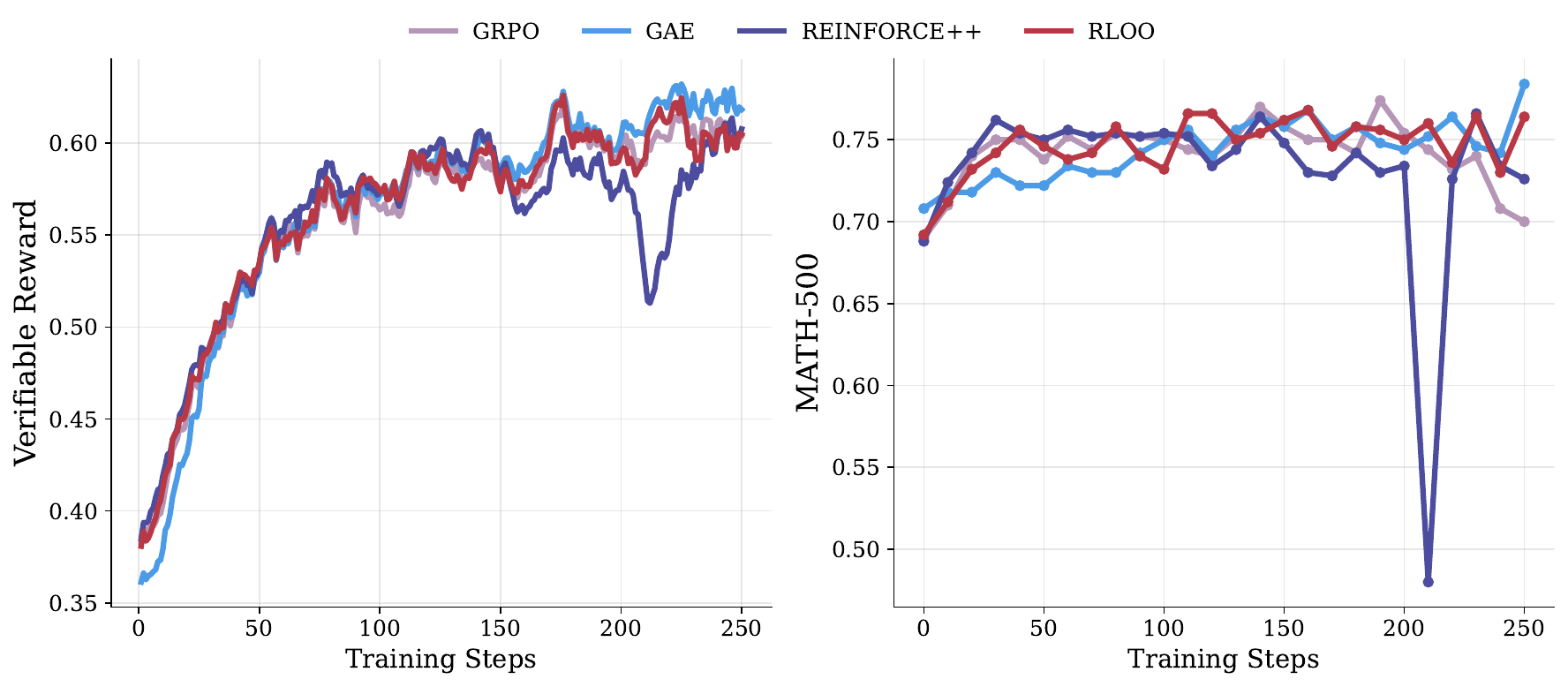}
    \caption{Training curves on Qwen2.5-7B using PURE-PRM+VR with different advantage estimators.}
    \label{fig:different_rl_algos}
\end{figure}
\subsection{PURE with Other RL Algorithms} \label{sec:other rl algo}
In this section, we apply PURE with various advantage estimators beyond RLOO, including GAE, GRPO, and REINFORCE++. The training curves are shown in Figure \ref{fig:different_rl_algos}. To make different algorithms compatible with the compound of verifier rewards and process rewards, we make adjustments similar to those in Eq. (\ref{eq:good adv}). For GRPO, the advantage is defined as:
\begin{equation}
    A_{i,t}=\frac{r^v_i - \text{mean}(\boldsymbol{r}^v)}{\text{std}(\boldsymbol{r}^v)} + \frac{\sum_{j=t}^{N}\gamma^{j-t} r^{p*}_{i,j} - \text{mean}\left(\sum_{l=1}^N\sum_{j=l}^N {\gamma^{j-l} \cdot \boldsymbol{r}^{p*}_{j}}\right)}{\text{std}\left(\sum_{l=1}^N\sum_{j=l}^N {\gamma^{j-l} \cdot \boldsymbol{r}^{p*}_{j}}\right)}
\end{equation}
where mean and standard deviation (std) are calculated over $K$ responses in a group. For REINFORCE++, the advantage is:
\begin{equation}
    A_{i,t}=r^v_i + \sum_{j=t}^{N}\gamma^{j-t} r^{p*}_{i,j}
\end{equation}

From Figure \ref{fig:different_rl_algos}, we find that all methods show similar performance before step 150. However, REINFORCE++ experiences a spike at around step 220. Although GAE converges a bit slower than others, it performs slightly more stably in the last 50 steps. Due to its additional learnable value network, GAE takes about 30\% more time for forward and backward passes. Considering performance, training time, and stability, we select RLOO as our preferred advantage estimator.

% \section{Future Work}
% There are several promising directions for further improving PRM-based RFT. First, developing generative PRMs is both urgent and crucial. As discussed in \S~\ref{sec:analysis}, current PRMs are unable to address the third type of reward hacking mentioned in \S~\ref{sec:reward hacking}, and also cannot evaluate the quality of patterns like endless repetition. Generative PRMs, however, could potentially resolve these issues by making better use of the strong language capabilities of LLMs. Second, iterative training between PRM and LLM is essential to ensure that the PRM continuously adapts to the output distribution of LLMs. We believe these areas offer valuable opportunities for future exploration.

\newpage
\section*{NeurIPS Paper Checklist}
\begin{enumerate}

\item {\bf Claims}
    \item[] Question: Do the main claims made in the abstract and introduction accurately reflect the paper's contributions and scope?
    \item[] Answer: \answerYes{} % Replace by \answerYes{}, \answerNo{}, or \answerNA{}.
    \item[] Justification: See abstract, section \ref{sec:intro} and \ref{sec:experiment}.

\item {\bf Limitations}
    \item[] Question: Does the paper discuss the limitations of the work performed by the authors?
    \item[] Answer: \answerYes{} % Replace by \answerYes{}, \answerNo{}, or \answerNA{}.
    \item[] Justification: See section \ref{sec:conclusion}.

\item {\bf Theory assumptions and proofs}
    \item[] Question: For each theoretical result, does the paper provide the full set of assumptions and a complete (and correct) proof?
    \item[] Answer: \answerNA{} % Replace by \answerYes{}, \answerNo{}, or \answerNA{}.
    \item[] Justification: The paper does not include theoretical results.

    \item {\bf Experimental result reproducibility}
    \item[] Question: Does the paper fully disclose all the information needed to reproduce the main experimental results of the paper to the extent that it affects the main claims and/or conclusions of the paper (regardless of whether the code and data are provided or not)?
    \item[] Answer: \answerYes{} % Replace by \answerYes{}, \answerNo{}, or \answerNA{}.
    \item[] Justification: See section \ref{sec:experiment}, appendix, and supplementary material.

\item {\bf Open access to data and code}
    \item[] Question: Does the paper provide open access to the data and code, with sufficient instructions to faithfully reproduce the main experimental results, as described in supplemental material?
    \item[] Answer: \answerYes{} % Replace by \answerYes{}, \answerNo{}, or \answerNA{}.
    \item[] Justification: See supplementary material.

\item {\bf Experimental setting/details}
    \item[] Question: Does the paper specify all the training and test details (e.g., data splits, hyperparameters, how they were chosen, type of optimizer, etc.) necessary to understand the results?
    \item[] Answer: \answerYes{} % Replace by \answerYes{}, \answerNo{}, or \answerNA{}.
    \item[] Justification: See section \ref{sec:experiment} and appendix.

\item {\bf Experiment statistical significance}
    \item[] Question: Does the paper report error bars suitably and correctly defined or other appropriate information about the statistical significance of the experiments?
    \item[] Answer: \answerNo{} % Replace by \answerYes{}, \answerNo{}, or \answerNA{}.
    \item[] Justification: All methods are tested using greedy decoding. Thus given a checkpoint, its benchmark scores are deterministic.

\item {\bf Experiments compute resources}
    \item[] Question: For each experiment, does the paper provide sufficient information on the computer resources (type of compute workers, memory, time of execution) needed to reproduce the experiments?
    \item[] Answer: \answerYes{} % Replace by \answerYes{}, \answerNo{}, or \answerNA{}.
    \item[] Justification: See section \ref{sec:experiment} and appendix.
    
\item {\bf Code of ethics}
    \item[] Question: Does the research conducted in the paper conform, in every respect, with the NeurIPS Code of Ethics \url{https://neurips.cc/public/EthicsGuidelines}?
    \item[] Answer: \answerYes{} % Replace by \answerYes{}, \answerNo{}, or \answerNA{}.
    \item[] Justification: The paper complies with the NeurIPS Code of Ethics.

\item {\bf Broader impacts}
    \item[] Question: Does the paper discuss both potential positive societal impacts and negative societal impacts of the work performed?
    \item[] Answer: \answerNA{} % Replace by \answerYes{}, \answerNo{}, or \answerNA{}.
    \item[] Justification: There is no societal impact of the work performed.
    
\item {\bf Safeguards}
    \item[] Question: Does the paper describe safeguards that have been put in place for responsible release of data or models that have a high risk for misuse (e.g., pretrained language models, image generators, or scraped datasets)?
    \item[] Answer: \answerNA{} % Replace by \answerYes{}, \answerNo{}, or \answerNA{}.
    \item[] Justification: The paper poses no such risks.

\item {\bf Licenses for existing assets}
    \item[] Question: Are the creators or original owners of assets (e.g., code, data, models), used in the paper, properly credited and are the license and terms of use explicitly mentioned and properly respected?
    \item[] Answer: \answerYes{} % Replace by \answerYes{}, \answerNo{}, or \answerNA{}.
    \item[] Justification: Both the base models from Qwen and dataset are properly credited.

\item {\bf New assets}
    \item[] Question: Are new assets introduced in the paper well documented and is the documentation provided alongside the assets?
    \item[] Answer: \answerNA{} % Replace by \answerYes{}, \answerNo{}, or \answerNA{}.
    \item[] Justification: The paper does not release new assets.

\item {\bf Crowdsourcing and research with human subjects}
    \item[] Question: For crowdsourcing experiments and research with human subjects, does the paper include the full text of instructions given to participants and screenshots, if applicable, as well as details about compensation (if any)? 
    \item[] Answer: \answerNA{} % Replace by \answerYes{}, \answerNo{}, or \answerNA{}.
    \item[] Justification: The paper does not involve crowdsourcing nor research with human subjects.

\item {\bf Institutional review board (IRB) approvals or equivalent for research with human subjects}
    \item[] Question: Does the paper describe potential risks incurred by study participants, whether such risks were disclosed to the subjects, and whether Institutional Review Board (IRB) approvals (or an equivalent approval/review based on the requirements of your country or institution) were obtained?
    \item[] Answer: \answerNA{} % Replace by \answerYes{}, \answerNo{}, or \answerNA{}.
    \item[] Justification: The paper does not involve crowdsourcing nor research with human subjects.

\item {\bf Declaration of LLM usage}
    \item[] Question: Does the paper describe the usage of LLMs if it is an important, original, or non-standard component of the core methods in this research? Note that if the LLM is used only for writing, editing, or formatting purposes and does not impact the core methodology, scientific rigorousness, or originality of the research, declaration is not required.
    %this research? 
    \item[] Answer: \answerNA{} % Replace by \answerYes{}, \answerNo{}, or \answerNA{}.
    \item[] Justification: The core method development in this research does not involve LLMs.

\end{enumerate}

\end{document}